\documentclass[conference]{IEEEtran}
\IEEEoverridecommandlockouts
\usepackage{cite}
\usepackage{amsmath,amssymb,amsfonts,bm}
\usepackage{url}
\usepackage{algorithmic}
\usepackage{graphicx}
\usepackage{booktabs}
\usepackage{multirow}
\usepackage{textcomp}
\usepackage{xcolor}
\usepackage[utf8]{inputenc}
\def\BibTeX{{\rm B\kern-.05em{\sc i\kern-.025em b}\kern-.08em
    T\kern-.1667em\lower.7ex\hbox{E}\kern-.125emX}}
\begin{document}

% Title
\title{Classification of Keratitis from Eye Corneal Photographs using Deep Learning\thanks{This work was financed by National Funds through the Portuguese funding agency, Portuguese Foundation for Science and Technology (FCT) through the Ph.D. Grant ``2020.06434.BD''.}}

% Authors
\author{\IEEEauthorblockN{Maria Miguel Beirão}
\IEEEauthorblockA{\textit{University of Porto}\\
\textit{INESC TEC}\\
Porto, Portugal\\
maria.m.beirao@inesctec.pt}
\and
\IEEEauthorblockN{João Matos}
\IEEEauthorblockA{\textit{MIT}\\
Cambridge, MA, USA\\
jcmatos@mit.edu }
\and
\IEEEauthorblockN{Tiago Gonçalves}
\IEEEauthorblockA{\textit{University of Porto} \\
\textit{INESC TEC}\\
Porto, Portugal \\
tiago.f.goncalves@inesctec.pt}
\and
\IEEEauthorblockN{Camila Kase}
\IEEEauthorblockA{\textit{Universidade Federal de São Paulo}\\
São Paulo, Brazil \\
camila.kase@unifesp.br}
\and
\IEEEauthorblockN{Luis Filipe Nakayama}
\IEEEauthorblockA{\textit{MIT}\\ {\textit{Universidade Federal de São Paulo}} \\
Cambridge, MA, USA \\
luis.naka@mit.edu}
\and
\IEEEauthorblockN{Denise de Freitas}
\IEEEauthorblockA{\textit{Universidade Federal de São Paulo}\\
São Paulo, Brazil \\
dfreitas.epm.unifesp@gmail.com}
\and
\IEEEauthorblockN{Jaime S. Cardoso}
\IEEEauthorblockA{\textit{University of Porto} \\
\textit{INESC TEC}\\
Porto, Portugal \\
jsc@fe.up.pt}
}

\maketitle

\begin{abstract}
Keratitis is an inflammatory corneal condition responsible for 10\% of visual impairment in low- and middle-income countries (LMICs), with bacteria, fungi, or amoeba as the most common infection etiologies. While an accurate and timely diagnosis is crucial for the selected treatment and the patients' sight outcomes, due to the high cost and limited availability of laboratory diagnostics in LMICs, diagnosis is often made by clinical observation alone, despite its lower accuracy. In this study, we investigate and compare different deep learning approaches to diagnose the source of infection: 1) three separate binary models for infection type predictions; 2) a multitask model with a shared backbone and three parallel classification layers (Multitask V1); and, 3) a multitask model with a shared backbone and a multi-head classification layer (Multitask V2). We used a private Brazilian cornea dataset to conduct the empirical evaluation. We achieved the best results with Multitask V2, with an area under the receiver operating characteristic curve (AUROC) confidence intervals of 0.7413-0.7740 (bacteria), 0.8395-0.8725 (fungi), and 0.9448-0.9616 (amoeba). A statistical analysis of the impact of patient features on models' performance revealed that sex significantly affects amoeba infection prediction, and age seems to affect fungi and bacteria predictions.

\end{abstract}


% Introduction
\section{Introduction}
According to the World Health Organization (WHO) \cite{who2023blindness}, corneal blindness impacts 6M people worldwide, with \textit{keratitis} as its leading cause~\cite{Ting2021,WHO2024}, with different consequences in low, middle and high-income countries~\cite{Austin2017}. Keratitis (or corneal ulcer) is an inflammation of the cornea caused by microorganisms (e.g., bacteria (BK), viruses (VK), fungi (FK), amoeba (AK), or herpes simplex keratitis (HSK)) or by non-infectious factors (e.g., trauma or chemical exposure)~\cite{keratitis}. This disease is more common in LMICs, with its risk factors (e.g., compromised immune systems, poor ocular hygiene, contact lens wear, eye trauma, and surgery, exposure to contaminated water and soils and agricultural work)~\cite{JohnsHopkinsMedicine,PreventBlindness} being correlated with the context of rural areas of these geographies~\cite{Ting2021}. The symptoms of keratitis include eye redness, eye pain, excess tears or discharge, foreign body sensation, difficulty opening the eyelid, blurred vision, light sensitivity, and watery eyes~\cite{JohnsHopkinsMedicine}. While the symptoms are well-defined, the main challenge with this disease is related to the diagnosis of the etiology, a crucial detail for the treatment of the condition and prevention of vision loss. The clinical pipeline for determining the etiology includes techniques such as corneal scrapping with gram stain, microscopy, and culture analysis. However, given their substantial costs (i.e., approximately \$200 per patient in Brazil), these resources are often unavailable in LMICs. The alternative approach relies on performing a clinical examination with a slit lamp, with bright light and enough magnification to observe the infection. Nevertheless, distinguishing between the types of infection remains difficult and might lead to a misdiagnosis, worsening the symptoms and requiring an eye surgery~\cite{taylor2009, Dalmon2012,Austin2017}. In this paper, we explore the potential of deep learning to classify keratitis etiologies using eye corneal photographs, making the following scientific contributions:
\begin{enumerate}
    \item Proposal and implementation of deep learning algorithms to classify different etiologies of keratitis from corneal eye photographs, using a new private Brazilian cornea dataset with clinical attributes and annotations;

    \item Assessment of the statistical impact of correlated features (i.e., age and sex) on disease presentation.
\end{enumerate}
The code related to the implementation of this work is publicly available in a GitHub repository\footnote{\url{https://github.com/mariamiguel01/Keratitis_Classifier}}.

% Related Work
\section{Related Work}
\subsection{Literature Review on Keratitis Classification}
In this section, we present a literature review on the state-of-the-art deep learning algorithms applied to keratitis classification in external eye photographs and organize our findings divided by predictive task: a) \textit{BK versus FK}, and b) \textit{Multiple Infections}.
\paragraph{BK versus FK} Reed et al.~\cite{Redd2022} developed a solution that uses a data augmentation pipeline and regularization techniques to fine-tune different convolutional neural network (CNN) architectures. After hyper-parameter optimization, their best model achieved an AUROC of 83\%. In another study conducted on a different Indian dataset comprising 50\% BK and 50\% FK, Reed et al. \cite{Redd2021} leveraged transfer learning on a pre-trained deep learning system and reported an accuracy of 76\%.

\paragraph{Multiple Infections}
Koyama et al.~\cite{Koyama2021} developed a deep learning algorithm to classify BK, FK, AK, and HSK using pre-trained CNNs and a dataset containing 4,306 images from clinical cases and web sources under different illumination conditions. Under a k-fold cross-validation strategy, they trained a hierarchical two-step CNN classifier to handle class imbalance, with the first CNN discriminating BK vs. non-bacterial cases and the second categorizing non-bacterial cases into AL, FK, or HSK. The final classifier, a gradient-boosting decision tree (GBDT), uses the CNN scores.

\subsection{Ophthalmologists versus Artificial Intelligence}
Different studies have been conducted to evaluate the ophthalmologist's performance metrics when compared to those of deep learning solutions. Human performance is often lower than that of the deep learning system and much more dependent on other variables like years of expertise and familiarity with the disease. 
Since this performance is very dependent on the dataset, several studies compared the model and the experts' performance. Redd et al.~\cite{Redd2022} compared the performance of 66 international expert cornea specialists in the same Indian dataset used in~\cite{Redd2021}. The panel of specialists achieved an AUROC of 76\%, which is significantly lower ($p<0.01$) than the one found by the deep learning model (83\%).

\section{Methodology}
This study was conducted in accordance with the principles of the Declaration of Helsinki and was approved by the Research Ethics Committee of the Federal University of São Paulo (UNIFESP), Brazil.

\subsection{Dataset}
\subsubsection{Description}
\begin{figure}[thbp]
    \centering
    \begin{minipage}{0.15\textwidth}
        \centering
        \includegraphics[width=\linewidth]{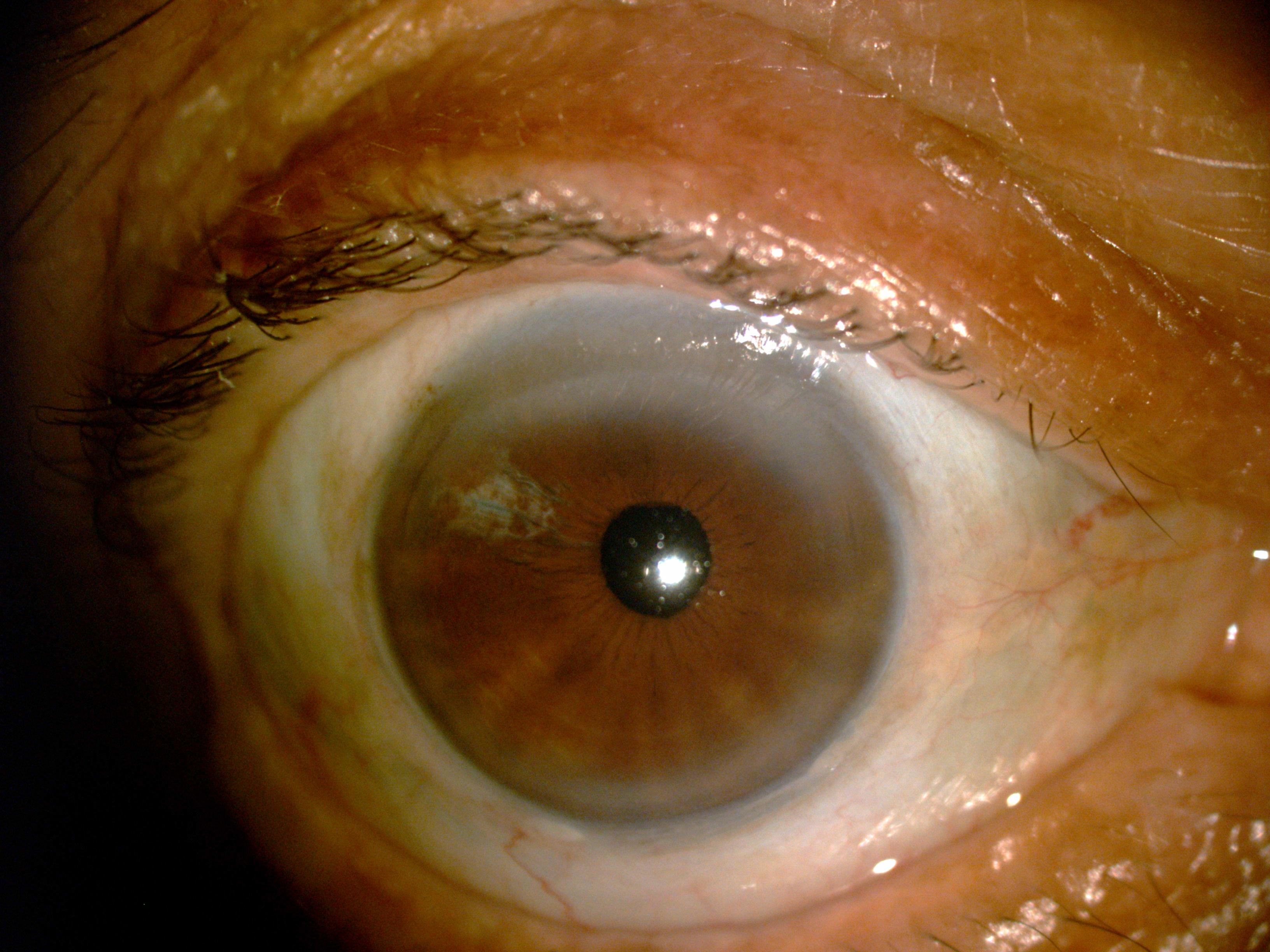}
        \caption{Bacteria Dataset Example.}
        \label{fig:ex1}
    \end{minipage}\hfill%
    \begin{minipage}{0.15\textwidth}
        \centering
        \includegraphics[width=\linewidth]{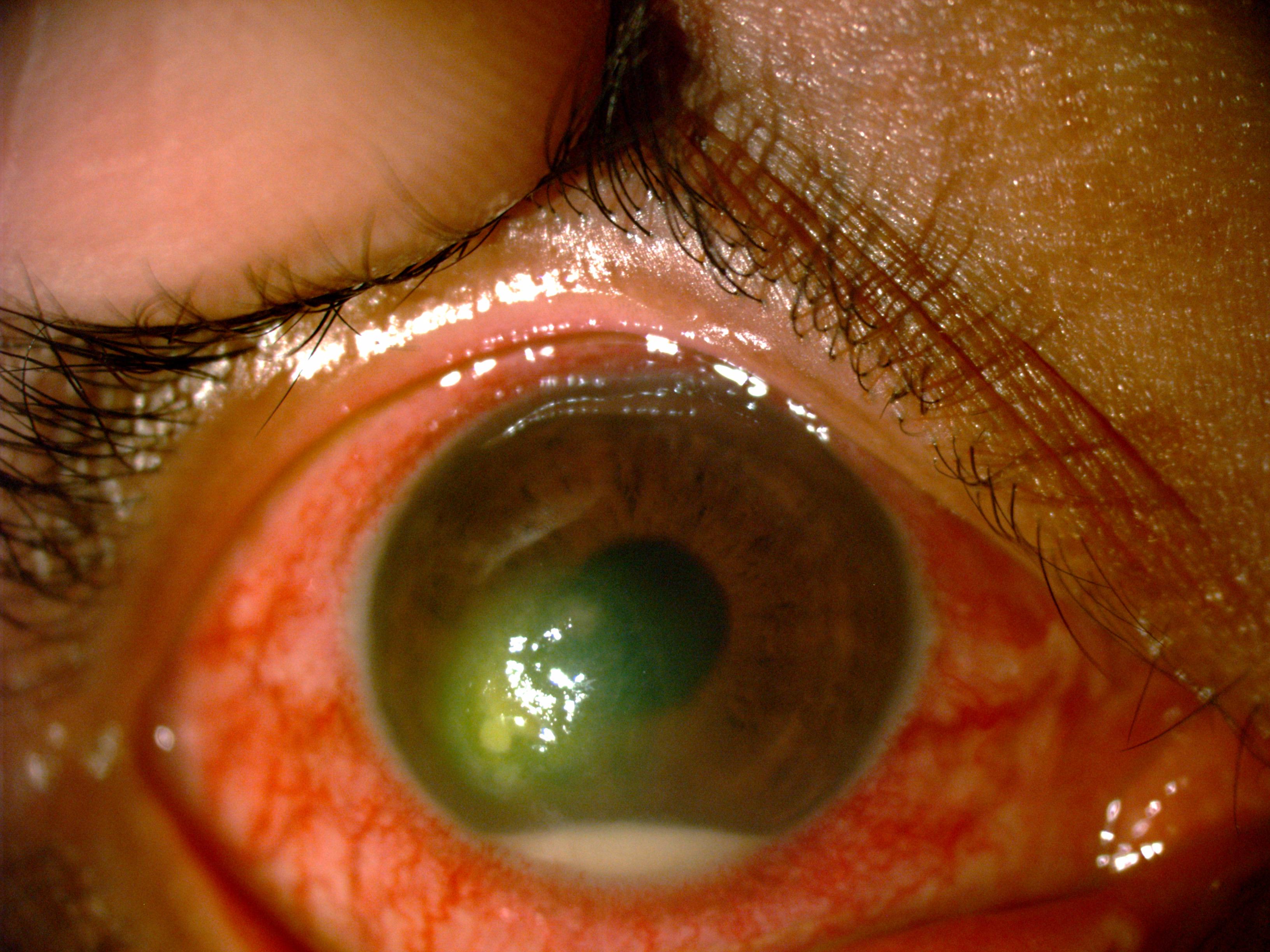}
        \caption{Fungi Dataset Example.}
        \label{fig:ex3}
    \end{minipage}\hfill%
    \begin{minipage}{0.15\textwidth}
        \centering
        \includegraphics[width=\linewidth]{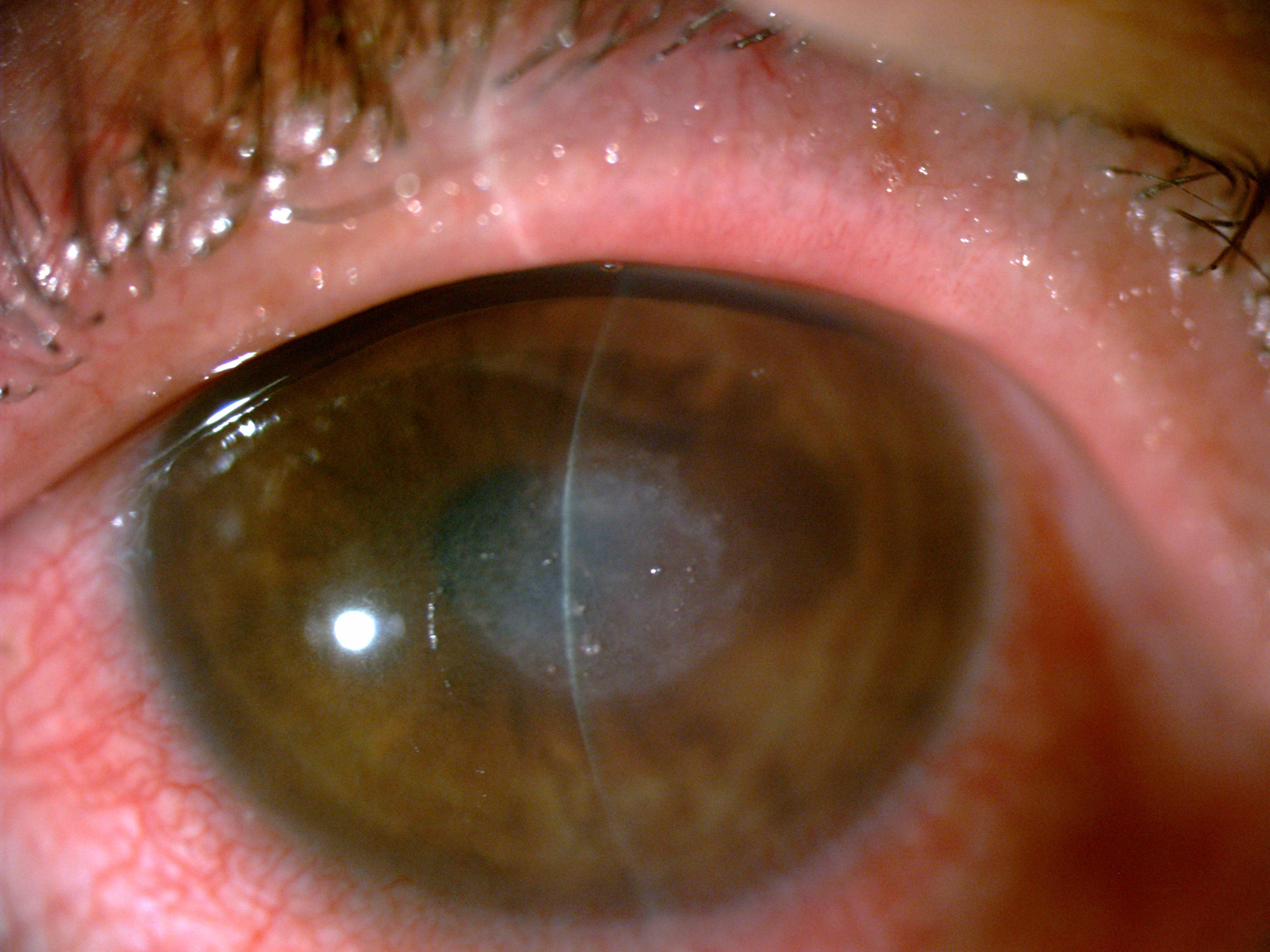}
        \caption{Amoeba Dataset Example.}
        \label{fig:ex5}
    \end{minipage}
    \vspace{0.5cm} % Add some vertical space between rows
    \begin{minipage}{0.15\textwidth}
        \centering
        \includegraphics[width=\linewidth]{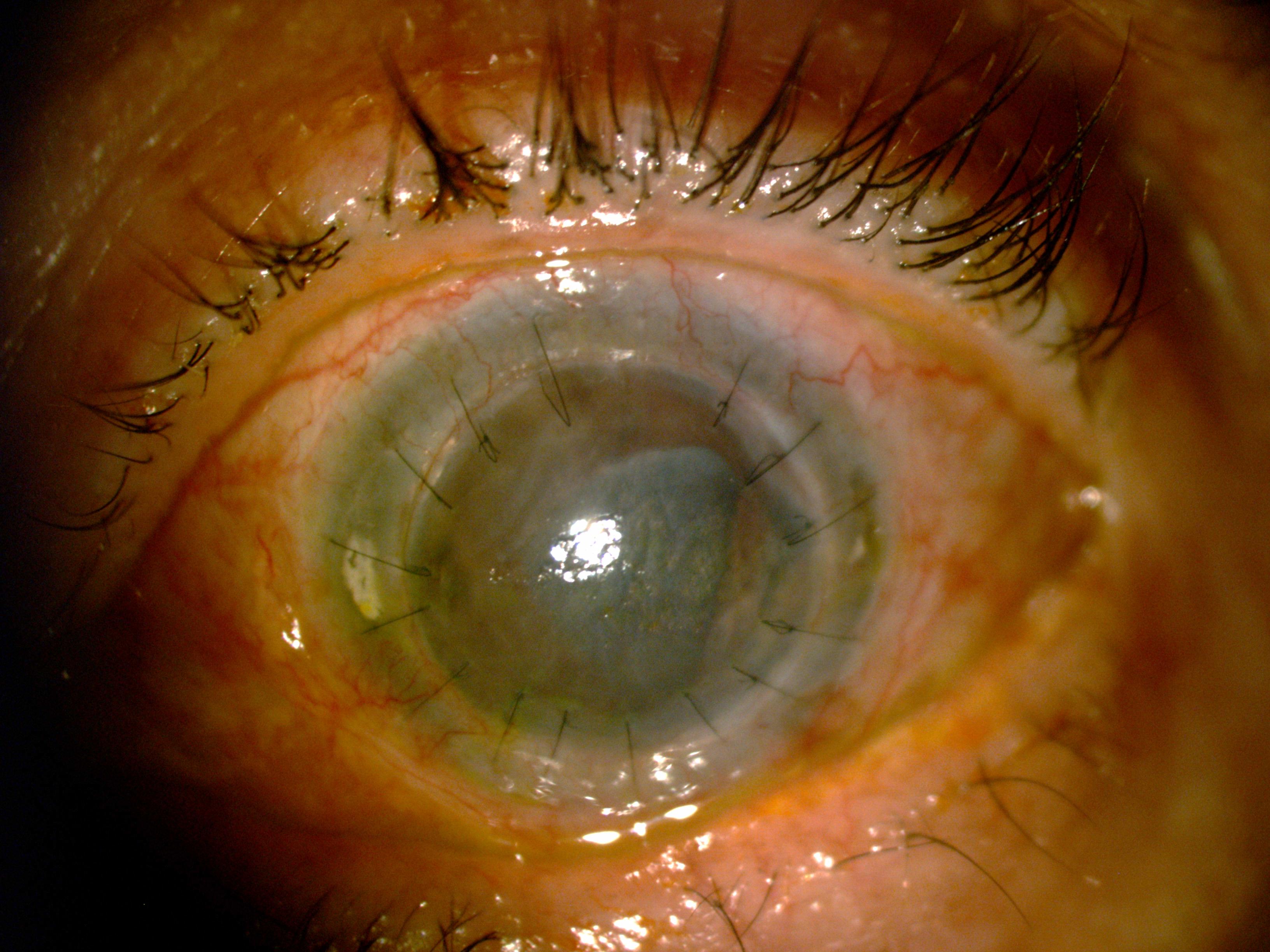}
        \caption{Bacteria Dataset Example.}
        \label{fig:ex2}
    \end{minipage}\hfill%
    \begin{minipage}{0.15\textwidth}
        \centering
        \includegraphics[width=\linewidth]{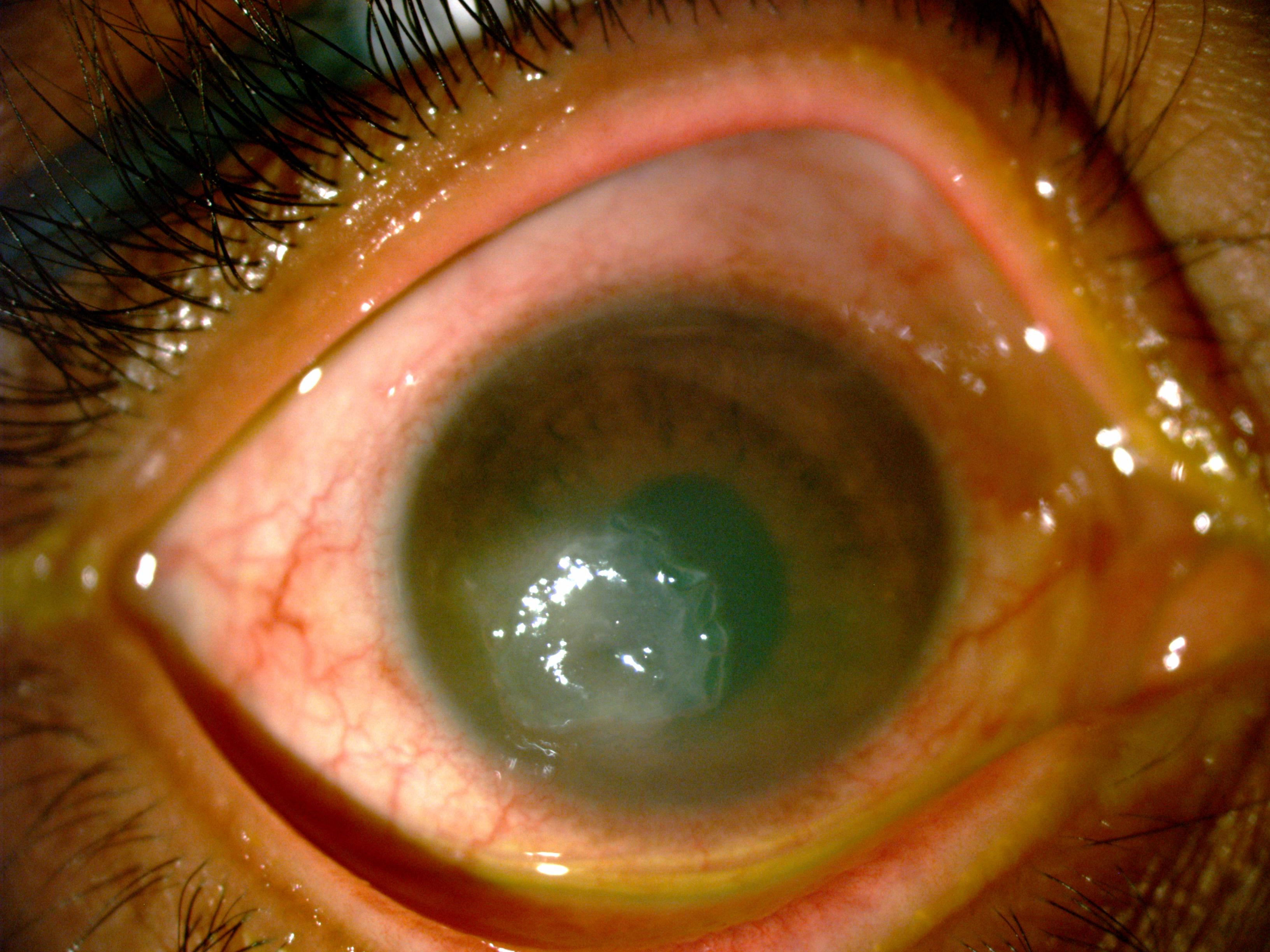}
        \caption{Fungi Dataset Example.}
        \label{fig:ex4}
    \end{minipage}\hfill%
    \begin{minipage}{0.15\textwidth}
        \centering
        \includegraphics[width=\linewidth]{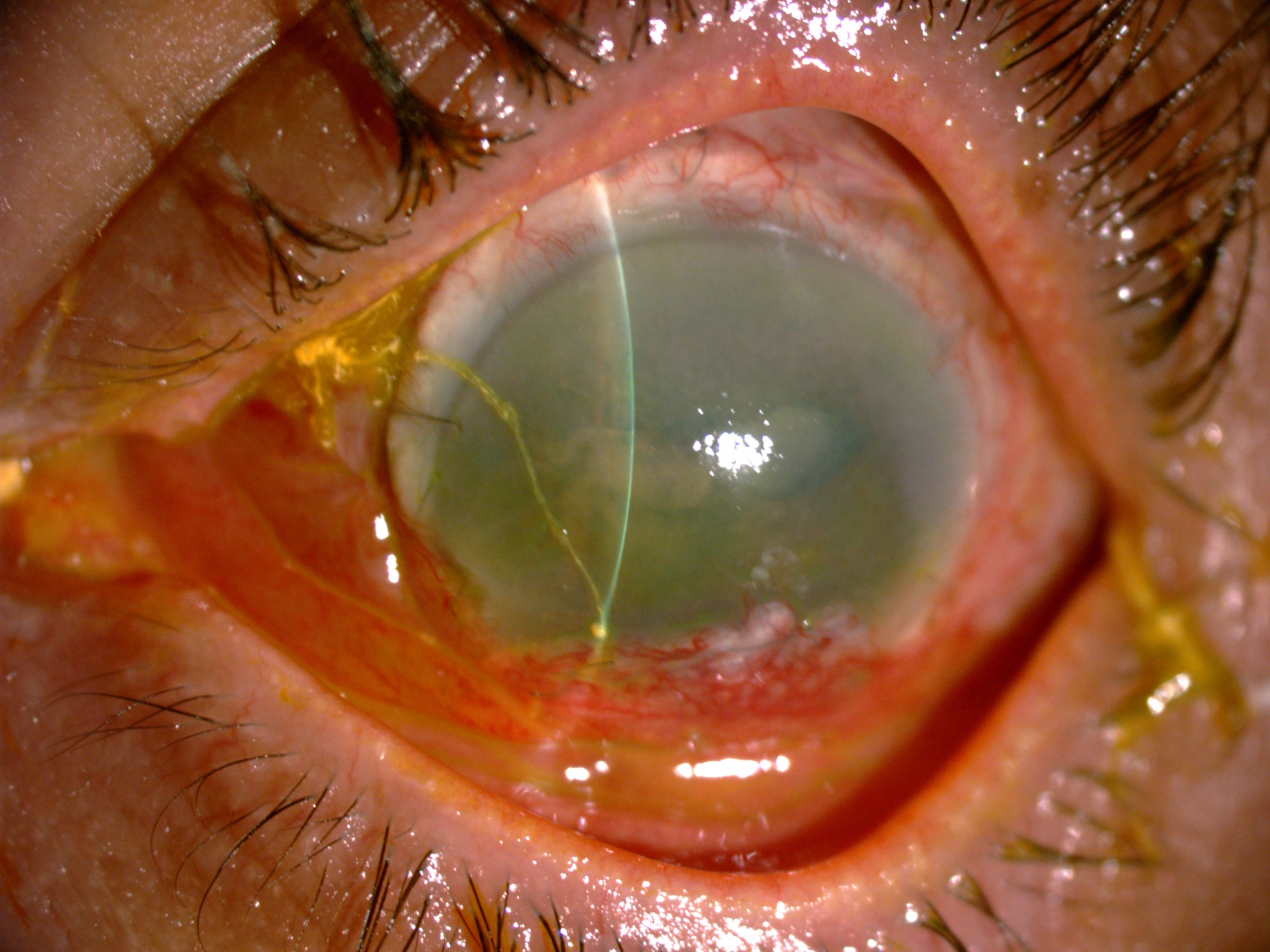}
        \caption{Amoeba Dataset Example.}
        \label{fig:ex6}
    \end{minipage}
\end{figure}

Two ophthalmologist co-authors (Camila Kase and Luis Nakayama) acquired the data at UNIFESP. This dataset contains healthy and non-healthy patients for keratitis (i.e., AK, BK, and FK), confirmed by corneal scraping culture. The dataset contains 24,692 exams, belonging to 3,296 patients (i.e., multiple exams per patient), associated with the patient's clinical data and the corresponding eye corneal images. From the total exams, we used the subset that had medical information and tested positive to keratitis (4,767 exams). After validation with the clinicians, we realized that the same image corresponded to different patient visits, and that an image is labeled as positive for an infection if any test for that infection is positive. These details allowed us to clean the dataset from redundant or non-relevant entries, and achieve a final dataset of 2,585 unique cases. We also excluded the cases with no infection (512 cases), resulting in a final dataset of 2,064 cases, the proportions of each class are present in Table ~\ref{tab:statisticasdata}. Figures~\ref{fig:ex1} -~\ref{fig:ex6} show some examples of the dataset.

\begin{table}[htbp]
\centering
\caption{Number and proportion of cases for the possible combinations of infections (Absent combinations did not occur in the dataset).}
\label{tab:statisticasdata}
\begin{tabular}{@{}ccccc@{}}
\toprule
\textbf{Bacteria} & \textbf{Amoeba} & \textbf{Fungi} & \textbf{\# Cases} & \textbf{\% Cases} \\ \midrule
Pos. & Neg. & Neg. & 1176 & 56.98\% \\
Neg. & Neg. & Pos. & 277 & 13.42\% \\
Pos. & Neg. & Pos. & 213 & 10.32\% \\
Neg. & Pos. & Neg. & 207 & 10.03\% \\
Pos. & Pos. & Neg. & 191 & 9.26\% \\ \bottomrule
\end{tabular}
\end{table}

\subsubsection{Exploratory Data Analysis}
Following clinical advice, we focus our exploratory data analysis on \textit{age} and \textit{sex}. We analyzed the age distribution using a kernel density estimation, observing a high peek of ages 20-40 and 40-65, which can be related to the two most common sources of infection (i.e., inadequate use of contact lenses and agricultural work). 
For implementation purposes, we grouped age values into four bins: $0-18$ (label 0), $18-40$ (label 1), $40-65$ (label 2), and $>65$ (label 3). The majority of cases exist in label 2 (39.39\%) and label 1 (36.05\%), while label 0 is underrepresented (2.62\%). Regarding sex, our analysis showed that there are 1,203 \textit{male} patients (58,28\%) and 861 \textit{female} patients (41,72\%) (i.e., balanced dataset for sex). We also performed a correlation analysis between the clinical features and the etiologies to understand the intrinsic biases of the dataset. Results showed positive and negative correlations between the etiologies and the clinical features (e.g., sex or age). Naturally, it is important to denote that there is a high correlation (both positive and negative) between the different types of infections.

\subsection{Implementation}
\subsubsection{Models}
\begin{figure}[th]
  \begin{center}
    \leavevmode
    \includegraphics[width=\columnwidth]{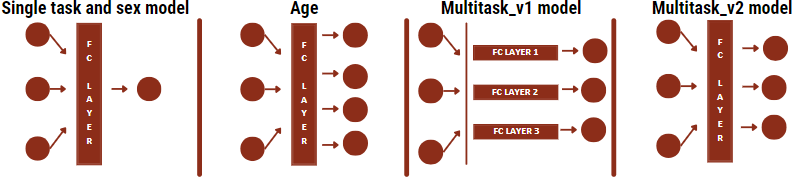}
    \caption{Schematic overview of the fully connected layers of each model, from left to right: \textbf{Baseline approach} and \textbf{Sex prediction} - a fully connected layer with a single neuron (and a sigmoid activation function);  \textbf{Age prediction} - a fully connected layer with four neurons (and a softmax activation function); \textbf{Multi-task V1}: - three parallel fully connected layers, each layer with a single neuron (and a sigmoid activation function), i.e., each layer predicting the binary outcome of a given infection; \textbf{Multi-task V2} - a fully connected layer with three neurons (and a sigmoid activation function), i.e., each neuron predicting the binary outcome of a given infection.}
    \label{fig:fcs}
  \end{center}
\end{figure}

We propose and implement two different approaches to achieve deep learning models to predict keratitis from eye cornea photographs:
\begin{itemize}
    \item{\textbf{Baseline approach}}: three simple binary deep learning models, one for each type of infection. Diagnosis for all infections requires the use of the three models.
    
    \item{\textbf{Multi-task approach}}: a multi-label deep learning model that predicts all the infections in a single forward pass, reducing runtime and computational costs. Within this approach, we tested two different architectures: \textit{1) Multi-task V1}, containing three binary classification layers, one per infection type; and \textit{2) Multi-task V2}, containing a classification layer with three neurons.
\end{itemize}
Besides, following the correlation analysis between clinical features and the clinical outcomes, we also developed models to predict age (multi-class model) and sex (binary model) from the same images to understand if we could mitigate potential biases~\cite{predicting}. To select an appropriate backbone (i.e.,  the feature extractor) for the models, we trained three candidates (i.e., DenseNet121~\cite{Huang_2017_CVPR}, ResNet50~\cite{He_2016_CVPR} and VGG16~\cite{simonyan2015vgg}) in a small subset (i.e., 50 images) of the dataset and recorded their runtime and performance. We selected the backbone with a lower runtime and better performance. After the backbone, we added a 2D average pooling layer with a kernel of 3$\times$3, stride of 2$\times$2, and padding of 1, followed by a batch normalization layer and a dropout layer with a probability of 0.3. Figure~\ref{fig:fcs} shows the classification layer of each of the approaches.

\subsubsection{Data Preparation and Augmentation}
Given the reduced number of usable cases (i.e., 2,064 out of 24,692), we started by artificially creating photographs of ``left'' and ``right'' eyes per patient, by applying a flip operation on the $y$-axis to the available images in the dataset (each patient only had an image of an eye). Besides increasing the size of the dataset, it also allows the model to be robust against images of both eyes. Afterwards, we resized all the images in the dataset to $224\times224$. We present the conversion of labels for the models in Table~\ref{tab:rules}.
\begin{table}[htbp]
\centering
\caption{Conversion of the clinical attributes into numerical labels.}
\label{tab:rules}
\begin{tabular}{@{}ll@{}}
\toprule
\multicolumn{1}{c}{\textbf{Clinical Attribute}} & \multicolumn{1}{c}{\textbf{Transformed (Original) Data}} \\ \midrule
Age & 0 (0-17), 1 (18-39), 2 (40-64), 3 ($>$65) \\
Infection Test Result & 0 (Negative), 1 (Positive) \\
Sex & 0 (Male), 1 (Female) \\ \bottomrule
\end{tabular}
\end{table}

We applied online data augmentation during training. The pipeline employed in this work consisted of random rotations (up to 20 degrees), random vertical flips (with a probability of 0.5), Gaussian blur ($kernel=5\times 5$, $0.1<\sigma<2$) and random color transformations (i.e., brightness, contrast, saturation, and hue adjustments) with the same parameters described in the literature~\cite{hu2023}. We also applied a z-score normalization to the images using ImageNet's mean ($[0.485, 0.456, 0.406]$) and standard deviation ($[0.229, 0.224, 0.225]$).

\subsection{Training Details}
\subsubsection{K-fold Strategy}
We implemented a k-fold strategy ($k=10$) for all the experiments. For the single-task models, we preserved the class distribution in the splits, while for the multi-task (i.e., multi-label), we ensured that every class and combination of classes were properly balanced. 
For each fold, we retained 10\% of the dataset as the test set to validate the best model, and, we split the remaining dataset to achieve, 80\% for train and 10\% for validation. This division was done in the dataset without the flipped images, and those were later added to the subset where the non-flipped image was, to avoid data leakage.

\subsubsection{Training and Validation Pipeline}
We trained the models for a certain number of epochs (between 100 and 300) and defined the optimal case for each. We initialized the models with ImageNet pre-trained weights. We have frozen the backbone of the models during the first ten epochs. Regarding loss functions, we used the binary cross-entropy for the binary classification and multi-task (multi-label) models, and cross-entropy for the multi-class models (age). To mitigate the issues related to class imbalance, we used class weights in the loss functions for all models. We used the adaptive moment estimation (Adam) optimizer and defined the best hyper-parameters (e.g., learning rate and weight decay) for each model by monitoring the validation loss (i.e., the minimum). We saved the best model according to the best value in the validation loss and an ``early-stopping'' routine. We used a batch-size of 16 for all the models.

\subsubsection{Multi-task Optimization}
Apart from the details described above, we added a few improvements to the multi-task models. We complemented the binary cross-entropy loss with a new term that takes into account that the cost of misidentifying a high-cost infection should be higher than that of misidentifying a low-cost infection. Adding this term to the loss makes sense because keratitis is common in LMICs, meaning that an incorrect treatment plan will have severe financial impacts on the patient and the healthcare providers. The average cost of each infection was computed using the insights provided by our clinical partners. To do so, we multiplied the prices of the medicines used in each infection (i.e., R\$45.2 for bacterial keratitis, R\$203 for fungal keratitis, and R\$95.5 for amoeba keratitis) by the number of flasks of each medicine used and the maximum duration of the treatment in months. We then divided these values by the cost of all infections. The values obtained were considered the hospital costs. After that, the combined loss was calculated using Equation~\ref{eq:losses}:
\begin{equation}
    \mathcal{L}_{total} = 0.8 \cdot (C_{w} \cdot  \mathcal{L}_{BCE})+0.2 \cdot  (H_{w} \cdot \mathcal{L}_{BCE}), 
\label{eq:losses}
\end{equation}
where $\mathcal{L}_{total}$ is the total loss, $C_{w}$ is the class weights, $\mathcal{L}_{BCE}$ is the binary cross-entropy loss, and $H_{w}$ is the hospital weights (described above). Finally, we integrated an adaptive threshold solution to consider a certain probability of a prediction. In this case, the optimal threshold would maximise the true positive rate (TPR) and minimise the false positive rate (FPR). To find this threshold, we computed Youden's J for every threshold value, comprising the receiver operating characteristic (ROC) curves of each task. The threshold that maximised J was then considered the optimal classification threshold for that infection~\cite{brownlee2020thresholdclass}.

\subsection{Evaluation Details}
\subsubsection{Performance Metrics}
To evaluate the predictive performance of the models, we computed the accuracy (ACC), AUROC, balanced accuracy (BA), F1-score, mean absolute error (MAE), precision, and recall~\cite{bishop2006}. During the training phase, we used these metrics and the evolution of the loss (in both training and validation splits) to understand the evolution of the model. In the test set, we computed these metrics, along with the confusion matrices (CFs) and the F1-score and recall for each class (within the same label). For the multi-task models, we also generated the joint CF of all infections. Lastly, to understand the TPR and FPR, we plotted the ROC curves. To ensure and validate the models' diagnosis, we generated saliency maps for every fold and one example of each class's positive and negative case. 

\subsubsection{Statistical Significance}
Since we implemented a k-fold strategy, we computed the average for all these metrics (and confusion matrices) across folds. We computed a $95\%$ normal confidence interval (CI) for the results of each metric and task, across models. After that, and only on models related to keratitis classification, we studied the statistical significance of the sex and age for the diagnosis. The goal of this task was to understand how these parameters can impact the result and make proper conclusions on the need for disentanglement of patient features from the images before inputting them into the model based on that. To achieve that, for each task and metric, we grouped the age and sex subsets. For the sex analysis, since only two independent groups were being compared, we used the t-test, whereas, for the age analysis, Analysis of Variance (ANOVA) was used. To account for the family-wise error rate of conducting multiple tests on the same dataset, Holm-Bonferroni Correction was implemented~\cite{Haynes2013}.

\subsubsection{Patients' Information Subsets}
To evaluate the influence of the age group and also the sex in the prediction of the etiology, the F1-Score, recall, precision, AUROC, ACC, BA, and MAE were computed by sex and age bin, for each task, meaning that a comparison of performance could be done in between sexes and ages for a specific task and metric, allowing for a deeper understanding of the influence of this factors in the classification.

% Results and Discussion
\section{Results and Discussion}
We obtained the best results with the DenseNet121 backbone, trained for 200 epochs, using Adam optimiser with a learning rate of $1\times10^{-6}$, and a weight decay of $1\times10^{-8}$.

\subsection{Sex and Age Prediction}
\begin{figure}[thbp]
  \begin{center}
    \leavevmode
    \includegraphics[width=\columnwidth]{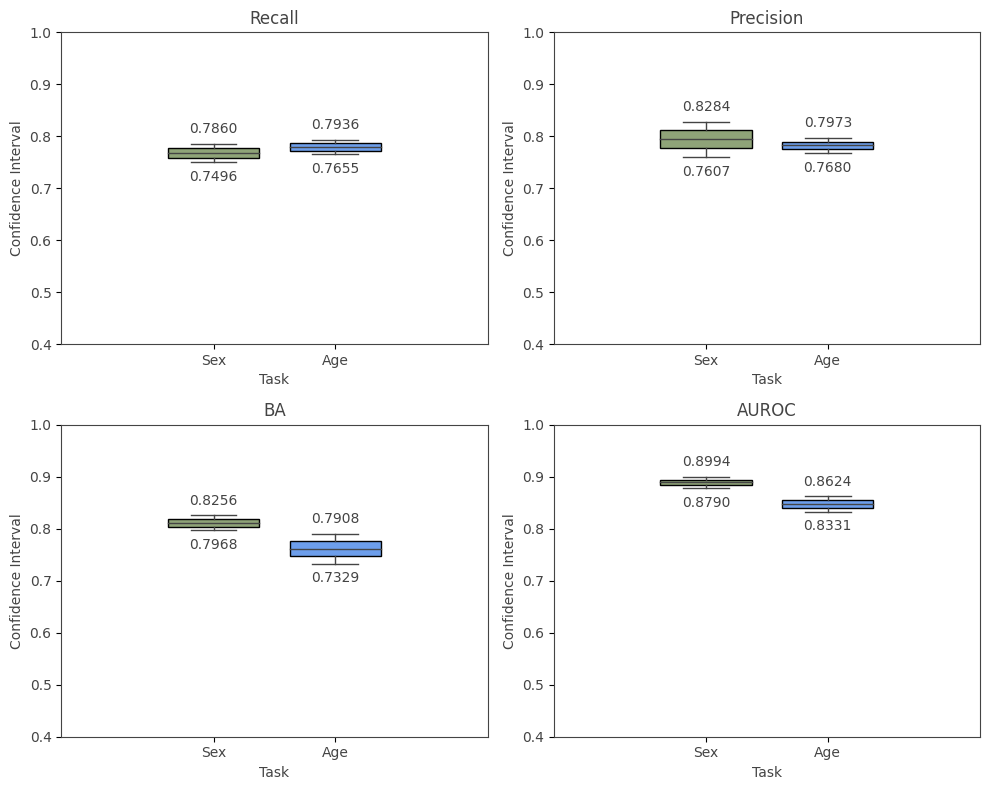}
        \caption{Results for the sex and age prediction models.}
    \label{fig:sexage}
  \end{center}
\end{figure}

For the tasks of sex and age prediction, our results suggest that the age bin is less predictable than the sex from the eye photograph (see Figure~\ref{fig:sexage}). On the other hand, unexpectedly, results for both tasks were fairly good, meaning that sex and age can be derived from eye corneal photos, and potentially influencing the prediction for keratitis, confirming the need for subset analysis. A look at the AUROCs, we can see that for sex the AUROC lies in $0.8790-0.8994$ and for age in $0.8331-0.8624$. For the MAE across folds, a CI of $0.2891-0.3114$ was found, which is low. This potentially means that when an incorrect classification happens, it does not occur for the most erroneous case. To validate this assumption, we looked into the CFs.

\subsection{Keratitis Classification}
\begin{figure}[th]
  \begin{center}
    \leavevmode
    \includegraphics[width=\columnwidth]{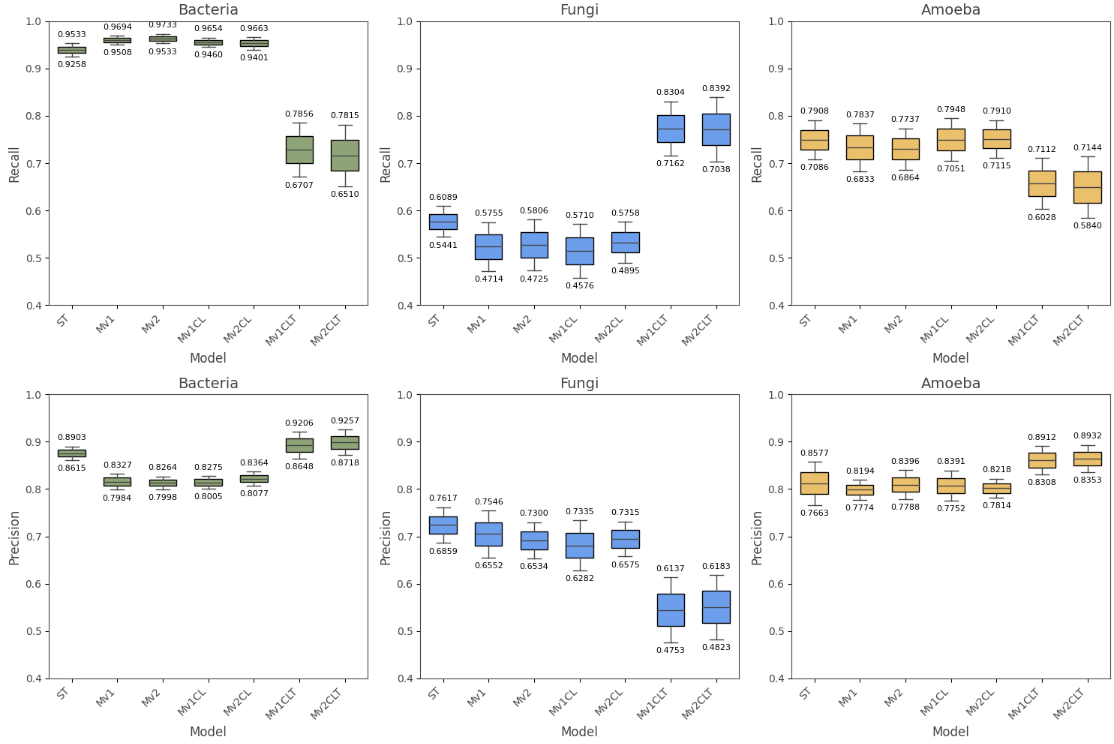}
    \caption{Recall and Precision of all models. Note: \textbf{ST}--Single task, \textbf{Mv1}--MultitaskV1, \textbf{Mv2}--MultitaskV2, \textbf{Mv1CL}--MultitaskV1 with clinical loss, \textbf{Mv2CL}--MultitaskV2 with clinical loss, \textbf{Mv1CLT}--MultitaskV1 with clinical loss and adaptive threshold, \textbf{Mv2CLT}--MultitaskV2 with clinical loss and adaptive threshold}
    \label{fig:all}
  \end{center}
\end{figure}

In the analysis of the results for the keratitis classification models, we focused on the precision and recall for all the etiologies because those were the metrics where the most differences were found. We excluded the baseline approach (single-task) from the pool of best models, since it needed to run three times to output a complete diagnosis, and did not use potential correlations between the etiologies to make the prediction, which could jeopardize the multi-infection case. From an attentive look at Figure~\ref{fig:all}, we realize that both Multi-task V1 and V2 models had the worst results for amoeba when compared to the models that had the optimal loss optimization. Besides, we observe an increase in incorrect classification (for all tasks) for the models that had the adaptive threshold, portrayed by a decrease of recall for bacteria and amoeba and a decrease in precision for fungi. Multi-task V1 and V2 with the clinical loss (see Equation~\ref{eq:losses}) attained very similar results, with the V2 being slightly better for recall in fungi and amoeba, and precision in bacteria.
This model obtained AUROCs of $0.9448-0.9617$ for amoeba,$ 0.8395-0.8725$ for fungi, and $0.7413-0.7740$ for bacteria. If we focus our attention on recall and precision, we observe that recall was especially relevant for amoeba and fungi, and precision for bacteria (possibly explained by the data imbalance). For instance, the low values found for fungi mean that there are several false negatives (FNs). This analysis is complemented with the CF (see Table~\ref{tab:cf_matrices_v2clinical}), which shows us that the model struggles with predicting the positive class correctly (once again, possibly due to data imbalance). Please note that we can derive similar conclusions for amoeba and bacteria with respect to the false positives (FPs). Looking at the extended CF (see Table~\ref{tab:cfsexpandedv2clinicaloss}), we can see that incorrect classification often happens because the model can not distinguish between a single or combined infection: for instance, from all the fungi infections, 19 were predicted as bacteria infections, 12 as fungi infections, and 21 as a combined infection of bacteria and fungi. These results support the need for more data.

\label{tab:cf_matrices_v2clinical}
\begin{table}[htbp]
\centering
\caption{Average CFs for the Multi-task V2 model trained with the clinical loss (see Equation~\ref{eq:losses}), for the different tasks.}

\begin{tabular}{@{}lccccc@{}}
\toprule
\textbf{Task} & \textbf{Actual/Predicted} & \textbf{Negative} & \textbf{Positive} \\ \midrule
\multirow{2}{*}{\textbf{Amoeba}} & Negative & 318.4 & 14.8 \\
 & Positive & 19.8 & 59.8 \\ \midrule
\multirow{2}{*}{\textbf{Bacteria}} & Negative & 31.1 & 65.7 \\
 & Positive & 14.8 & 301.2 \\ \midrule
\multirow{2}{*}{\textbf{Fungi}} & Negative & 292.0 & 22.8 \\
 & Positive & 45.8 & 52.2 \\ \bottomrule
\end{tabular}
\end{table}

\begin{table}[htbp]
\centering
\caption{Confusion matrix for Multi-task V2 model trained with the clinical loss. Note: \textbf{H}--healthy, \textbf{B}--bacteria, \textbf{F}--fungi, \textbf{A}--amoeba, \textbf{B,F}--combined infection of bacteria and fungi, \textbf{F,A}--combined infection of fungi and amoeba, \textbf{B,A}--combined infection of bacteria and amoeba, \textbf{B,F,A}--combined infection of bacteria, fungi and amoeba.}
\label{tab:cfsexpandedv2clinicaloss}
\begin{tabular}{@{}ccccccccc@{}}
\toprule
 \textbf{Label\textbackslash Pred}& \textbf{H} & \textbf{B} & \textbf{F} & \textbf{A} & \textbf{B,F} & \textbf{F,A} & \textbf{B,A} & \textbf{B,F,A} \\ \midrule
\textbf{H} & 0 & 0 & 0 & 0 & 0 & 0 & 0 & 0 \\
\textbf{B} & 4 & 193 & 2 & 2 & 24 & 0 & 9 & 0 \\
\textbf{F} & 0 & 19 & 12 & 0 & 21 & 0 & 2 & 0 \\
\textbf{A} & 4 & 13 & 1 & 11 & 0 & 0 & 11 & 0 \\
\textbf{B,F} & 0 & 20 & 0 & 0 & 24 & 0 & 0 & 0 \\
\textbf{F,A} & 0 & 0 & 0 & 0 & 0 & 0 & 0 & 0 \\
\textbf{B,A} & 1 & 5 & 0 & 2 & 0 & 0 & 30 & 0 \\
\textbf{B,F,A} & 0 & 0 & 0 & 0 & 0 & 0 & 0 & 0 \\ \bottomrule
\end{tabular}
\end{table}

\subsection{Influence of Patient Features}
We chose the best-performing methodology for the identity feature evaluation. Given the interesting performance of the sex and age prediction models (i.e., the model was detecting identity features), we sought to assess if the classification model was using these features to output its predictions (i.e., the introduction of bias).

\subsubsection{Influence of Sex} 
Table~\ref{tab:metrics_age_sex} presents the summary of the performance metrics for sex (i.e., p-values after correction). We can observe that sex has an impact on the ACC of the detection of fungi infection, with male sex having lower values. Regarding the detection of amoeba infection, we found a significant difference for F1-score, recall and BA. This can potentially mean that the model is using features related to the sex of the patient in the final predictions (especially for amoeba), meaning that a disentanglement of these features from the corneal eye photograph could boost the performance of the model.

\subsubsection{Impact of Age}
Table~\ref{tab:metrics_age_sex} presents the summary of the performance metrics for age (i.e., p-values after correction). We found significant differences for all metrics across all age groups, with the exception of BA (i.e., age features may play a role in the prediction). The analysis of the values for the detection of fungi infection reported the same as the detection of bacteria infection, with all metrics exhibiting significant differences across age groups. Regarding the detection of amoeba infection, we found significant value for ACC. Please note that we do not report AUROC values for this task because of the lack of data on at least one age bin. Moreover, given the class imbalance, we argue that feature disentanglement is needed for this case, although an evaluation with a bigger and balanced dataset should also be done.

\begin{table}[htbp]
\centering
\caption{Corrected p-values for Age and Sex by Infection Type.}
\label{tab:metrics_age_sex}
\begin{tabular}{@{}lcc|cc|cc@{}}
\toprule
\textbf{} & \multicolumn{2}{c|}{\textbf{Bacteria}} & \multicolumn{2}{c|}{\textbf{Fungi}} & \multicolumn{2}{c}{\textbf{Amoeba}} \\ \midrule
\textbf{Metric}& \textbf{Age} & \textbf{Sex} & \textbf{Age} & \textbf{Sex} & \textbf{Age} & \textbf{Sex} \\ \midrule
F1-score & \textbf{0.0001} & 1.000 & \textbf{7.70}\bm{$\times$}\textbf{10\textsuperscript{-16}} & 0.2510 & 0.4555 & \textbf{0.0035} \\
Recall & \textbf{0.0153} & 1.000 & \textbf{1.63}\bm{$\times$}\textbf{10\textsuperscript{-12}} & 0.7070 & 0.9099 & \textbf{0.0010} \\
Precision & \textbf{0.0029} & 1.000 & \textbf{2.93}\bm{$\times$}\textbf{10\textsuperscript{-20}} & 0.1320 & 0.4946 & 0.2580 \\
BA & 0.7693 & 1.000 & \textbf{1.24}\bm{$\times$}\textbf{10\textsuperscript{-13}} & 1.000 & 0.2418 & \textbf{0.0042} \\
ACC & \textbf{0.0006} & 1.000 & \textbf{6.83}\bm{$\times$}\textbf{10\textsuperscript{-16}} & \textbf{0.0003} & \textbf{0.0055} & 0.6840 \\
AUROC & - & 1.000 & - & 1.000 & - & 0.4810 \\
\bottomrule
\end{tabular}
\end{table}
An analysis of the results for the other frameworks revealed similar patterns, making clear the need for further investigation of both sex and age in a larger dataset.

% Conclusions and Future Work
\section{Conclusions and Future work}
Although the proposed solution has yielded good results, several improvements could be made in the future to reduce FPs for the detection of bacteria infection and FNs for the detection of fungi and amoeba infections. As explained earlier, we believe that the problem is caused by the lack of data on fungi and amoeba infections, causing the model to not predict these classes correctly. Since we do not train the model with healthy patients, when this happens, the eye exams are often classified as bacterial infections, causing FPs. Hence, the first step in future work would be to either acquire more data or artificially generate new images with generative adversarial networks (GANs) or diffusion models. Another interesting test would be to evaluate the performance of the model on segmented images of the cornea, as well as test other state of the art backbones. It would also be interesting to introduce the sex and age of that patient as auxiliary inputs to the prediction model and evaluate the performance. Lastly, the developed work has proven that it is possible to predict the sex and age of the patient based on an eye photograph. This could mean that those identity features are present in the images, introducing new biases to our methodology. The obtained p-values pointed in that direction. To avoid those connections, the disentanglement of identity features from the eye corneal photographs could be relevant.

In conclusion, in this work, we demonstrated the effectiveness and potential of deep learning solutions in advancing keratitis diagnosis and contributing to broader healthcare improvements. The proposed solution achieved state-of-the-art metrics for a multi-label diagnosis of bacterial keratitis, fungal keratitis, and amoeba keratitis, with $95\%$ confidence intervals for AUROC of $0.7413-0.7740$, $0.8395-0.8725$, and $0.9448-0.9617$, respectively. When looking at the sex and age performance, interesting AUROC values were found, $0.8790-
0.8994$ for sex and $0.8331-0.8624$ for age. A significant impact of sex for amoeba infections and age for bacteria and fungi infections was found, meaning a bias associated with patient features could be implied.

% References
\bibliographystyle{plain}
\bibliography{refs}

\end{document}